# Efficient Legendre moment computation for grey level images

G.Y. Yang[a], H.Z. Shu[a], G.N. Han[b], C. Toumoulin[c], L.M. Luo[a]

[a]*Laboratory of Image Science and Technology, Department of Biology and Medical Engineering , Southeast University, 210096, Nanjing, People's Republic of China*

[b]*IRMA, Université Louis Pasteur et C.N.R.S., 7, rue René-Descartes F, 67084 Strasbourg, France*

[c]*Laboratoire Traitement du Signal et de l'Image, INSERM U642– Université de Rennes 1, Campus de Beaulieu, 35042 Rennes Cedex, France*

The information about the corresponding author is:

Huazhong Shu, Ph. D

Laboratory of Image Science and Technology

Department of Biology and Medical Engineering

Southeast University

210096 Nanjing

People's Republic of China

Tel: 86-25-83794249

Fax: 86-25-83792698

E-mail: shu.list@seu.edu.cn

**Abstract** Legendre orthogonal moments have been widely used in the field of image analysis. Because their computation by a direct method is very time expensive, recent efforts have been devoted to the reduction of computational complexity. Nevertheless, the existing algorithms are mainly focused on binary images. We propose here a new fast method for computing the Legendre moments, which is not only suitable for binary images but also for grey levels. We first set up the recurrence formula of one-dimensional (1D) Legendre moments by using the recursive property of Legendre polynomials. As a result, the 1D Legendre moments of order $p$, $L_p = L_p(0)$, can be expressed as a linear combination of $L_{p-1}(1)$ and $L_{p-2}(0)$. Based on this relationship, the 1D Legendre moments $L_p(0)$ is thus obtained from the array of $L_1(a)$ and $L_0(a)$ where $a$ is an integer number less than $p$. To further decrease the computation complexity, an algorithm, in which no multiplication is required, is used to compute these quantities. The method is then extended to the calculation of the two-dimensional Legendre moments $L_{pq}$. We show that the proposed method is more efficient than the direct method.

## 1 Introduction

Since Hu introduced the moment invariants [1], moments and moment functions of image intensity values have been successfully and widely used in the field of image analysis, such as object recognition, object representation, edge detection [2]. Orthogonal moments (e.g. Legendre moment and Zernike moment) can be used to represent an image with the minimum amount of information redundancy [3]. Since the computation of orthogonal moments of a two-dimensional (2D) image by a direct method involves a significant amount of arithmetic operations, some fast algorithms have been developed to reduce the computational complexity. However, the existing methods for fast computation of Legendre moments are mainly focused on binary image [4-6]. Now the moments of a grey level image are also used in many applications, such as texture analysis [7], therefore, we propose a fast algorithm for computing the Legendre moments for grey level images. The principle is as follows. The recurrence formula of one-dimensional (1D) Legendre moments is firstly established by using the recursive property of Legendre polynomials. The 1D Legendre moments of order $p$, $L_p = L_p(0)$, is expressed as a linear combination of $L_{p-1}(1)$ and $L_{p-2}(0)$. Based on this relationship, the 1D Legendre moments $L_p(0)$ can thus be obtained from the array of $L_1(a)$ and $L_0(a)$ where $a$ is an integer number less than $p$. An algorithm based on a systolic array in which no multiplication is required is used to compute these quantities. We propose then an extension of the method to the 2D Legendre moment $L_{pq}$ computation.

The remainder of this paper is organized as follows. In Section 2, we first describe a new approach for computing 1D Legendre moments for a 1D signal, and then extend the method to the 2D Legendre moment calculation. Section 3 gives the detailed analysis of the computational complexity and some experimental results. Section 4 provides some concluding remarks.

## 2. Fast computation of 2D Legendre moments

The $(p+q)$th-order Legendre moment of an image with intensity function $f(x, y)$ is defined by

$$L_{pq} = \frac{(2p+1)(2q+1)}{4} \int_{-1}^{1}\int_{-1}^{1} P_p(x)P_q(y)f(x,y)dxdy \tag{1}$$

where $P_p(x)$ is $p$th order Legendre polynomial given by

$$P_p(x) = \frac{1}{2^p}\sum_{k=0}^{n/2}(-1)^k \frac{(2p-2k)!}{k!(p-k)!(p-2k)!} x^{p-2k}, x \in [-1,1]. \tag{2}$$

For a digital image of size $N \times N$, Eq. (1) can be approximated by

$$L_{pq} = \frac{(2p+1)(2q+1)}{(N-1)^2}\sum_{i=1}^{N}\sum_{j=1}^{N} P_p(x_i)P_q(y_j)f(x_i,y_j) \tag{3}$$

with $x_i = (2i - N - 1)/(N - 1)$, $y_j = (2j - N - 1)/(N - 1)$.

The Legendre polynomial obeys the following recursive relation

$$P_{p+1}(x) = \frac{2p+1}{p+1} xP_p(x) - \frac{p}{p+1} P_{p-1}(x), p \geq 1 \tag{4}$$

with $P_0(x) = 1$, $P_1(x) = x$.

In the following, we present an algorithm for the fast calculation of the 2D Legendre moment for grey level images. For the sake of simplicity, let us first consider the computation of the 1D Legendre moments.

For a 1D discrete signal $f(x_i)$, $1 \leq i \leq N$, the 1D Legendre moment is given by

$$L_p = \frac{2p+1}{N-1}\sum_{i=1}^{N} P_p(x_i) f(x_i) \tag{5}$$

Let us now introduce the following notation

$$L_p(a) = \frac{2p+1}{N-1}\sum_{i=1}^{N} x_i^a P_p(x_i) f(x_i) \tag{6}$$

It can be easily seen that $L_p = L_p(0)$. Thus, we turn to the fast computation of $L_p(a)$ in the following.

Substitution of Eq. (4) into Eq. (6) yields

$$\begin{aligned} L_p(a) &= \frac{2p+1}{N-1}\sum_{i=1}^{N} x_i^a \left[\frac{2p-1}{p} x_i P_{p-1}(x_i) - \frac{p-1}{p} P_{p-2}(x_i)\right] f(x_i) \\ &= \frac{2p+1}{p}\frac{2p-1}{N-1}\sum_{i=1}^{N} x_i^{a+1} P_{p-1}(x_i) f(x_i) - \frac{p-1}{p}\frac{2p+1}{2p-3}\frac{2p-3}{N-1}\sum_{i=1}^{N} x_i^a P_{p-2}(x_i) f(x_i) \end{aligned} \tag{7}$$

therefore, we have the following recurrence relation for $p \geq 2$

$$L_p(a) = \frac{2p+1}{p}\left[L_{p-1}(a+1) - \frac{p-1}{2p-3} L_{p-2}(a)\right] \tag{8}$$

with

$$L_0(a) = \frac{1}{N-1}\sum_{i=1}^{N} x_i^a f(x_i) = \frac{1}{N-1} G_N(a) \tag{9}$$

$$L_1(a) = \frac{3}{N-1}\sum_{i=1}^{N} x_i^{a+1} f(x_i) = \frac{3}{N-1} G_N(a+1) \tag{10}$$

$$G_N(a) = \sum_{i=1}^{N} x_i^a f(x_i) \tag{11}$$

The above discussion shows that the 1D Legendre moments $L_p = L_p(0)$, for $p \geq 2$, can be deduced from the values of $L_0(a)$ and $L_1(a)$ where $a$ is an integer less than $p$, $L_0(a)$ and $L_1(a)$ being obtained by $G_N(a)$. The calculation of Eq. (11) leads to distinguish two different cases: $N$ odd and $N$ even.

(1) $N = 2L + 1$

Since $x_i = (2i - N - 1)/(N - 1)$, we deduce from Eq. (11) that

$$G_{2L+1}(a) = \sum_{i=1}^{2L+1} \left( \frac{2i - 2L - 2}{2L} \right)^a f(x_i) = \frac{1}{L^a} \sum_{i=1}^{2L+1} (i - L - 1)^a f(x_i)$$
$$= \begin{cases} \frac{1}{L^a} \left[ -L^a f(x_1) - (L-1)^a f(x_2) - \ldots - f(x_L) \right. \\ \quad \left. + f(x_{L+2}) + 2^a f(x_{L+3}) + \ldots + L^a f(x_{2L+1}) \right], & a \text{ is odd} \\ \frac{1}{L^a} \left[ L^a f(x_1) + (L-1)^a f(x_2) + \ldots + f(x_L) \right. \\ \quad \left. + f(x_{L+2}) + 2^a f(x_{L+3}) + \ldots + L^a f(x_{2L+1}) \right], & a \text{ is even} \end{cases} \tag{12}$$

Eq. (12) can be rewritten as

$$G_{2L+1}(a) = \begin{cases} \frac{1}{L^a} \sum_{i=1}^{L} i^a g_1(x_i), & a \text{ is odd} \\ \frac{1}{L^a} \sum_{i=1}^{L} i^a g_2(x_i), & a \text{ is even} \end{cases} \tag{13}$$

with

$$g_1(x_i) = f(x_{L+i+1}) - f(x_{L-i+1}), \quad i = 1,2,\ldots,L \tag{14}$$

$$g_2(x_i) = f(x_{L+i+1}) + f(x_{L-i+1}), \quad i = 1,2,\ldots,L \tag{15}$$

(2) $N = 2L$

Eq. (11) becomes

$$G_{2L}(a) = \sum_{i=1}^{2L} \left( \frac{2i - 2L - 1}{2L - 1} \right)^a f(x_i) = \frac{1}{(2L-1)^a} \sum_{i=1}^{2L} (2i - 2L - 1)^a f(x_i)$$
$$= \begin{cases} \frac{1}{(2L-1)^a} \left[ -(2L-1)^a f(x_1) - (2L-3)^a f(x_2) - \ldots - f(x_L) \right. \\ \quad \left. + f(x_{L+1}) + 3^a f(x_{L+2}) + \ldots + (2L-1)^a f(x_{2L}) \right], & a \text{ is odd} \\ \frac{1}{(2L-1)^a} \left[ (2L-1)^a f(x_1) + (2L-3)^a f(x_2) + \ldots + f(x_L) \right. \\ \quad \left. + f(x_{L+1}) + 3^a f(x_{L+2}) + \ldots + (2L-1)^a f(x_{2L}) \right], & a \text{ is even} \end{cases} \tag{16}$$

or

$$G_{2L}(a)=\begin{cases}\dfrac{1}{(2L-1)^a}\displaystyle\sum_{i=1}^{L}(2i-1)^a g_3(x_i), & a \text{ is odd}\\ \dfrac{1}{(2L-1)^a}\displaystyle\sum_{i=1}^{L}(2i-1)^a g_4(x_i), & a \text{ is even}\end{cases} \tag{17}$$

with

$$g_3(x_i)=f(x_{L+i})-f(x_{L-i+1}),\quad i=1,2,\ldots,L \tag{18}$$

$$g_4(x_i)=f(x_{L+i})+f(x_{L-i+1}),\quad i=1,2,\ldots,L \tag{19}$$

We discuss, in the following two subsections, the way to efficiently calculate $G_N(a)$ from Eqs. (13) or (17), according to the different modalities of the 1D signal $f(x_i)$.

2.1. $f(x_i) = 1$, for $i = 1, 2, \ldots, N$.

In this case, Eqs. (13) and (17) become

$$G_{2L+1}(a)=\begin{cases}0, & a \text{ is odd}\\ \dfrac{2}{L^a}\displaystyle\sum_{i=1}^{L} i^a, & a \text{ is even}\end{cases} \tag{20}$$

$$G_{2L}(a)=\begin{cases}0, & a \text{ is odd}\\ \dfrac{2}{(2L-1)^a}\displaystyle\sum_{i=1}^{L}(2i-1)^a=\dfrac{2}{(2L-1)^a}\left(\sum_{i=1}^{2L} i^a-2^a\sum_{i=1}^{L} i^a\right), & a \text{ is even}\end{cases} \tag{21}$$

The above equations show that to obtain the values of $G_N(a)$, we only need to calculate the following summation

$$H_M(a)=\sum_{i=1}^{M} i^a \tag{22}$$

For the computation of Eq. (22), which is just the 1D geometric moment of order $a$ of a 'binary' signal, we use the formulae proposed by Spiliotis and Mertzios [8]

$$H_M(1)=\frac{M(M+1)}{2},\quad H_M(2)=\frac{M(M+1)(2M+1)}{6}$$
$$H_M(3)=\frac{M^2(M+1)^2}{4},\quad H_M(4)=\frac{M(M+1)(2M+1)(3M^2+3M+1)}{30} \tag{23}$$

and for $a \geq 4$, the recurrence formula

$$\binom{a+1}{1}H_M(1)+\binom{a+1}{2}H_M(2)+\ldots+\binom{a+1}{a}H_M(a)=(M+1)^{a+1}-(M+1) \tag{24}$$

where $\binom{i}{j}=\frac{i!}{j!(i-j)!}$ is a combination number.

2. 2. $f(x_i) \neq f(x_j)$ for some $i \neq j$

Eq. (17) can be written as

$$G_{2L}(a)=\begin{cases}\frac{1}{(2L-1)^a}\sum_{i=1}^{2L} i^a h_1(x_i), & a \text{ is odd}\\ \frac{1}{(2L-1)^a}\sum_{i=1}^{2L} i^a h_2(x_i), & a \text{ is even}\end{cases} \tag{25}$$

where

$$h_1(x_i)=\begin{cases} g_3(x_{(i+1)/2}), \text{if } i \text{ is odd}\\ 0, \text{ otherwise}\end{cases} \tag{26}$$

$$h_2(x_i)=\begin{cases} g_4(x_{(i+1)/2}), \text{if } i \text{ is odd}\\ 0, \text{ otherwise}\end{cases} \tag{27}$$

Here $g_3(x_i)$ and $g_4(x_i)$ are given by Eqs. (18) and (19), respectively.

It can be seen from Eqs. (13) and (25) that we need to calculate the summation of the form

$$S_M(a)=\sum_{i=1}^{M} i^a g(x_i) \tag{28}$$

Note that $S_M(a)$ is the 1D geometric moment of order $a$ of an arbitrary 1D signal.

Since many algorithms are available in the literature to speed up the computation of Eq. (28), we decided to choose the method proposed by Chan et al. [9]. Their method is able to efficiently compute the grey level image moments. It makes use of a systolic array for computing the moments in which no multiplication is required. We recently applied such a method to efficiently calculating the Zernike moments [10].

Thus, the 1D Legendre moments $L_p(0)$, for $0 \le p \le M$ ($M$ denoting the maximal order we want to calculate), can be efficiently obtained using the previously presented algorithm. Fig. 1 shows the computation order of $L_p(0)$ for $p$ varied from 0 to 5.

Let us now describe the method for fast computation of the 2D Legendre moments $L_{pq}$. The double summation in Eq. (3) can be split into the following separate form

$$\begin{aligned} L_{pq} &= \frac{(2p+1)(2q+1)}{(N-1)^2}\sum_{i=1}^{N}\sum_{j=1}^{N} P_p(x_i)P_q(y_j)f(x_i,y_j) \\ &= \frac{2p+1}{N-1}\sum_{i=1}^{N} P_p(x_i)\left(\frac{2q+1}{N-1}\sum_{j=1}^{N} P_q(y_j)f(x_i,y_j)\right) = \frac{2p+1}{N-1}\sum_{i=1}^{N} P_p(x_i)Y_{iq} \end{aligned} \tag{29}$$

where $Y_{iq}$ is the $q$th-order row moment of row $i$ given by

$$Y_{iq} = \frac{2q+1}{N-1}\sum_{j=1}^{N} P_p(y_j)f(x_i,y_j) \tag{30}$$

These equations show that the computation of 2D Legendre moments of grey level images can be decomposed into two steps. First, the 1D Legendre moments $Y_{iq}$, for $1 \le i \le N$ and $0 \le q \le M$, are computed by using the algorithm described in Subsections 2.1 and 2.2, according to the different image modalities of $f(x_i, y_j)$. Then, the row moments $Y_{iq}$ are applied to compute the 2D Legendre moments $L_{pq}$. That is, after the first step, the 2D Legendre moments $L_{pq}$ can be calculated as a 1D moments by setting the image intensity function $f(x_i, y_j)$ to the $Y_{iq}$ previously computed. The algorithm for computing the 2D Legendre moments is depicted in figure 2. It should be pointed out that such a strategy can also be realized in parallel.

## 3. Computation complexity and experimental results

Let the image size be $N \times N$ pixels, and $M$ the maximum order of Legendre moments to be calculated. The direct computation of Eq. (3) requires approximately $M^2N^2/2$ additions and multiplications, respectively. In the following, we give a detailed analysis of the computational complexity of our algorithm.

We first briefly describe the computational complexity of the algorithm when applied to binary images. From Ref. [8], the computation of the geometrical moments up to the order $M$ of a binary image with $N \times N$ pixels, requires approximately $4M$ power calculations, $2M^2$ multiplications, and $M^2$ additions (note that these numbers are not dependent on $N$). The computation of the 2D Legendre moments $L_{pq}$, by using the recursive algorithm, needs $O(NM^3)$ additions and $O(M^3)$ multiplications. Therefore, the algorithm is very efficient in comparison with the direct method.

The computational complexity of the method for grey level images takes into account the parity of $N$.

(1) For odd value of $N$

Let us first consider the number of operations required in the computation of $i$th row moments $Y_{iq}$ ($0 \le q \le M$). Note that the functions $g_1(x)$ and $g_2(x)$ defined by Eqs. (14) and (15) are used for odd value of $N$. To obtain the values of $Y_{iq}$, we must calculate $G_N(a)$ with Eq. (13) for $0 \le a \le M$. This step needs only $(M+1)^2(N/2–1)$ additions. The computation of $Y_{iq}$ (for $0 \le q \le M$) from the pre-calculated $G_N(a)$, requires $M(M–1)/2$ additions and $2M(M–1)$ multiplications. Therefore, the computation of $N$ rows $Y_{iq}$ (for $1 \le i \le N$) needs approximately $M^2(N^2+N)/2$ additions and $2M^2N$ multiplications.

When all $Y_{iq}$, for $1 \le i \le$ N and $0 \le q \le M$, are obtained, the 2D Legendre moments $L_{pq}$, for $0 \le p+q \le M$, can be calculated in a similar way. The corresponding addition and multiplication numbers are $M^3N/12+M^2N$ and $2M^3/3+2M^2$.

Accordingly, the overall computation makes use of $M^2N^2/2+M^3N/12$ additions and $2M^2N+2M^3/3$ multiplications approximately.

(2) For even value of $N$

The functions $h_1(x)$ and $h_2(x)$, that are defined by Eqs. (26) and (27), will be used in the computation of $G_N(a)$. The only difference between case (2) and case (1) is that Eq. (25) is adopted instead of (13). The computation of Eq. (25) requires twice additions more than that is needed in Eq. (13). Thus, the total computational complexity is approximately $M^2N^2+M^3N/6$ additions and $2M^2N+2M^3/3$ multiplications.

The computational complexities of the proposed algorithm and the direct method are summarized in Table 1. The displayed values show that the new algorithm uses a much fewer number of multiplications than the direct method, what consequently leads to a more efficient computation time.

We illustrate the algorithm efficiency with two binary and grey level images of size 256×256, respectively (first column of Table 2). The first one displays a Chinese character while the second one shows some butterflies. In both cases, the reconstruction of the image was performed using the following relation

$$\hat{f}(x_i, y_j) = \sum_{p=0}^{M} \sum_{q=0}^{p} L_{p-q,q} P_{p-q}(x_i) P_q(y_j) \tag{31}$$

The results are provided for different orders of the moments, and are shown in Table 2.

## 5. Conclusion

In this paper, a new fast algorithm for computing the 2D Legendre moments of grey level images has been presented. The proposed method has the following advantages:

(1) The 1D Legendre moments can be obtained by a recurrence relation. Moreover, the initial value used in the iterative method can be calculated with additions only.

(2) The 2D moment computation can be decomposed into two 1D moment calculation.

(3) It does not require so many multiplications that as the direct method, leading thus to a better efficiency in terms of computational time.

(4) The algorithm can be implemented in parallel.

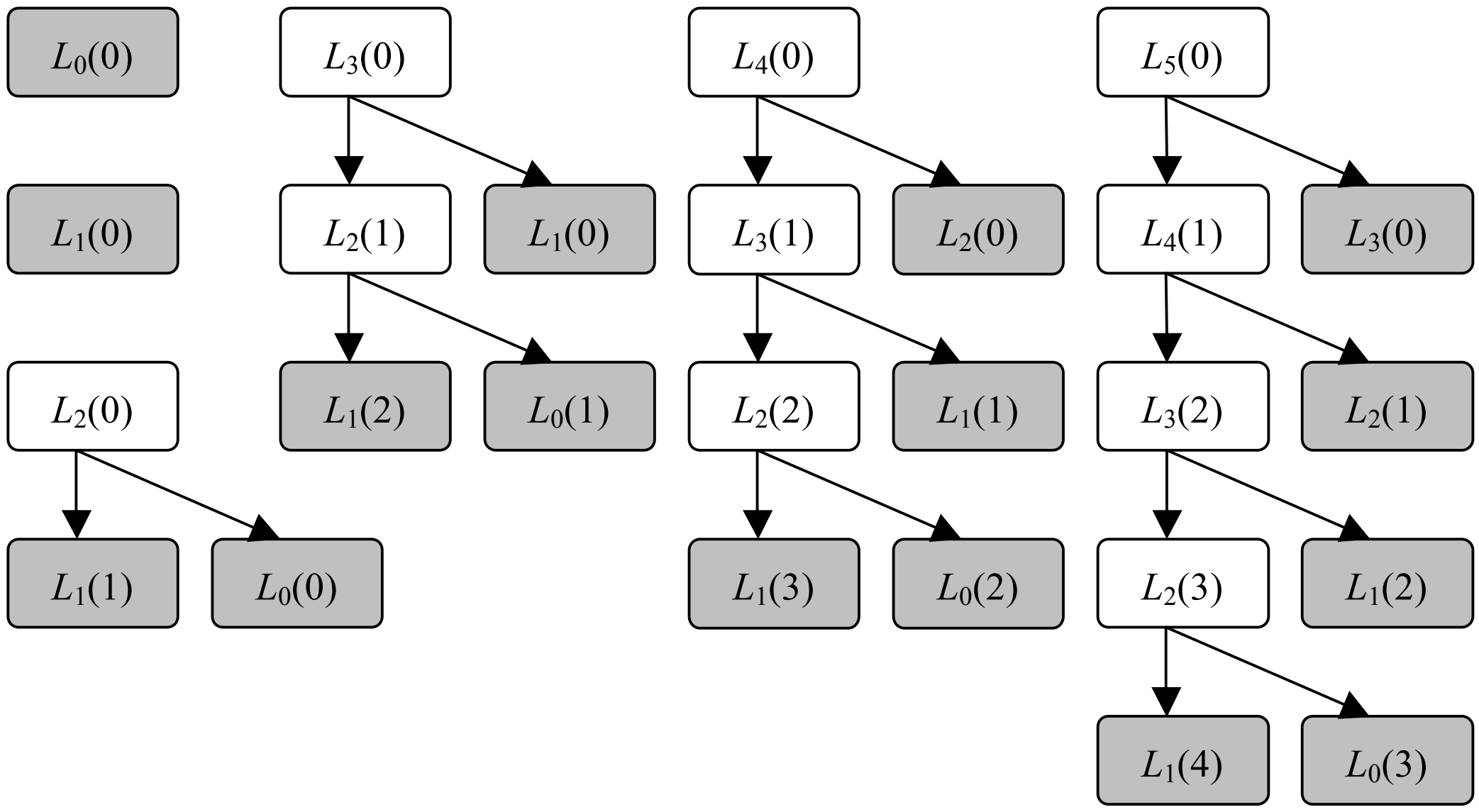

Fig. 1. Computation process of $L_p(0)$ with $p$ from 0 to 5. Grey level boxes correspond to already computed coefficients and white boxes to coefficients that will be computed from those which appear in grey level boxes.

for $i = 1$ to $N$

computing $L_0(q)$ ($0 \le q \le M$-2) and $L_1(q)$ ($0 \le q \le M$-1) using Eqs. (9) and (10)

for $q = 0$ to $M$

computing $Y_{iq}$ for each row $i$ of the image using Eq. (8)

endfor

endfor

for $p = 0$ to $M$

computing $L_0(a)$ ($0 \le a \le M$-$p$-2) and $L_1(a)$ ($0 \le a \le M$-$p$-1) using Eqs. (9) and (10) from pre-calculated $Y_{iq}$

for $q = 0$ to $M - p$

computing the 2D Legendre moments $L_{pq}$ using recursive method

endfor

endfor

Fig. 2. Algorithm for computing $L_{pq}$

| | | Number of additions | Number of multiplications |
|---|---|---|---|
| $N$=256 $M$=40 | direct method | $\approx M^2N^2/2$<br>56 426 496 | $\approx M^2N^2/2$<br>56 426 496 |
| | our method | $\approx M^2N^2+M^3N/6$<br>109 735 680 | $\approx 2NM^2+2M^3/3$<br>798 720 |
| $N$=255 $M$=40 | direct method | $\approx M^2N^2/2$<br>55 986 525 | $\approx M^2N^2/2$<br>55 986 525 |
| | our method | $\approx M^2N^2/2+M^3N/12$<br>54 010 530 | $\approx 2NM^2+2M^3/3$<br>795 600 |

Table 1 Comparison of the computational complexity for the two methods

| original | $M=24$ | $M=32$ | $M=40$ |
|---|---|---|---|
| 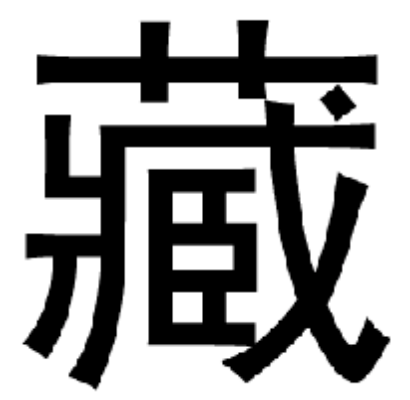 | 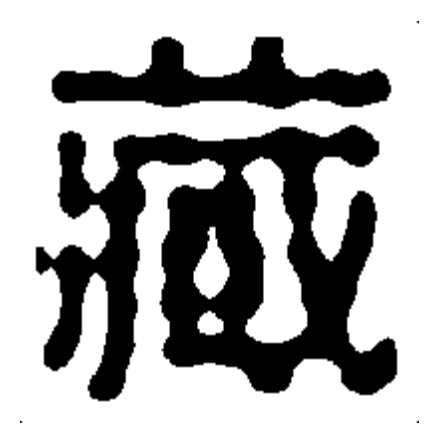 | 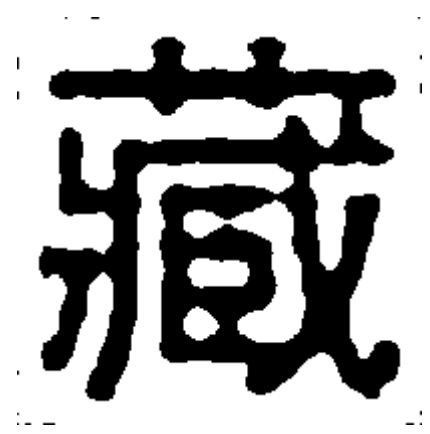 | 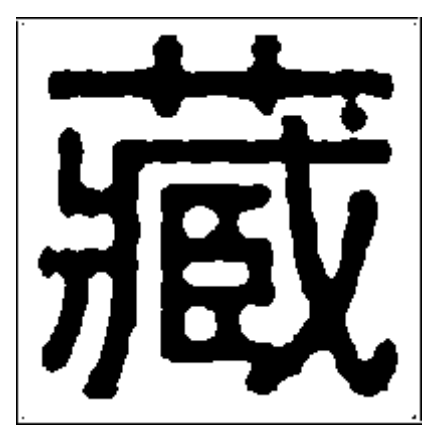 |
| 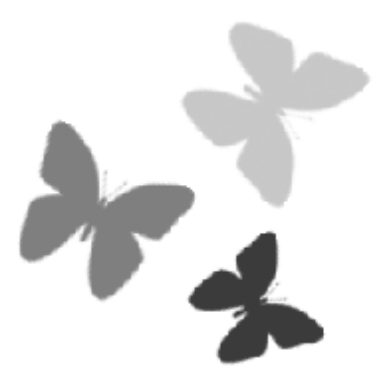 | 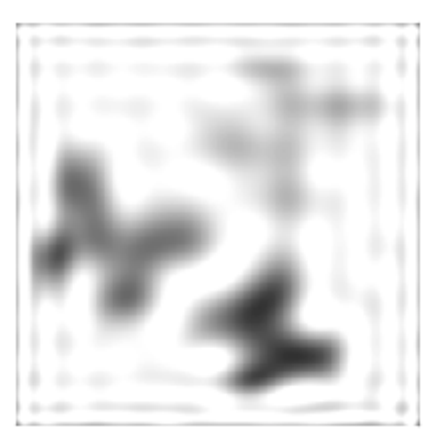 | 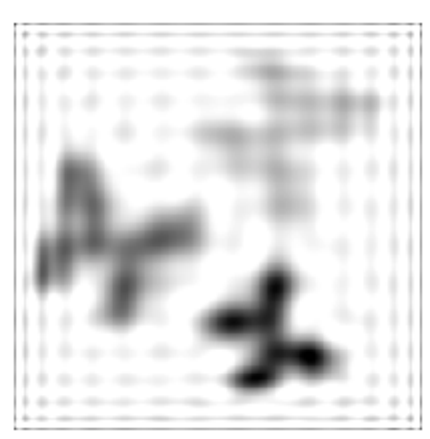 | 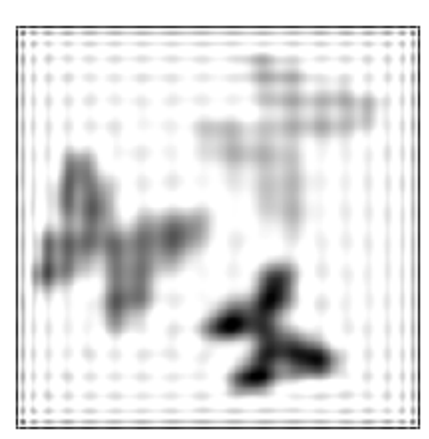 |

Table 2 Results of reconstruction image for different order 24, 32, 40

**About the Author**—GUANGYU YANG received the B. S. Degree in Biology and Medical Engineering from Southeast University, China, in 2002. He is now a graduate student of the Department of Biology and Medical Engineering of Southeast University. His current research is mainly focused on image processing and virtual reality.

**About the Author**—HUAZHONG SHU received the B. S. Degree in Applied Mathematics from Wuhan University, China, in 1987, and a Ph. D. degree in Numerical Analysis from the University of Rennes (France) in 1992. He was a postdoctoral fellow with the Department of Biology and Medical Engineering, Southeast University, from 1995 to 1997. His recent work concentrates on the treatment planning optimization, medical imaging, and pattern recognition.

**About the Author**—GUONIU HAN received the B. S. Degree in Applied Mathematics from Wuhan University, China, in 1987, and a Ph. D. degree in Mathematics from the University of Strasbourg I (France) in 1992. He works as Research Associate (CR) at French National Center for Scientific Research (CNRS) since 1993. His recent work focuses on the algebraic combinatorics, computer algebra and pattern recognition.

**About the Author**—CHRISTINE TOUMOULIN, PhD, is Associate Professor at the Technological Institute of the University of Rennes 1. She has been involving in Image Processing with special emphasis on biomedical imaging since 1987. Her contribution deals with 2-D and 3-D image segmentation, pattern recognition as well as registration techniques. She has served as Editorial Assistant for the *IEEE Transactions on Biomedical Engineering* journal from July 1996 to June 2001. Since 2002, she is an associate editor for the same journal.

**About the Author**—LIMIN LUO obtained his Ph. D. degree in 1986 from the University of Rennes (France). Now he is a professor of the Department of Biology and Medical Engineering, Southeast University, Nanjing, China. He is the author and co-author of more than 80 papers. His current research interests include medical imaging, image analysis, computer-assisted systems for diagnosis and therapy in medicine, and computer vision. Dr LUO is a senior member of the IEEE. He is an associate editor of *IEEE Eng. Med. Biol. Magazine* and *Innovation et Technologie en Biologie et Medecine* (ITBM).